\def\year{2020}\relax
\documentclass[letterpaper]{article} % DO NOT CHANGE THIS
\usepackage{aaai20}  % DO NOT CHANGE THIS
\usepackage{times}  % DO NOT CHANGE THIS
\usepackage{helvet} % DO NOT CHANGE THIS
\usepackage{courier}  % DO NOT CHANGE THIS
\usepackage[hyphens]{url}  % DO NOT CHANGE THIS
\usepackage{graphicx} % DO NOT CHANGE THIS
\usepackage{graphicx}  % DO NOT CHANGE THIS
\usepackage{epsfig}
\usepackage{amsmath}
\usepackage{amssymb}
\usepackage[utf8]{inputenc}
\usepackage{amsfonts,bm,color,soul}
\usepackage{amsopn}
\usepackage{amssymb}
\usepackage{empheq}
\usepackage{epstopdf}
\usepackage{float}
\usepackage{framed}

\newtheorem{lemma}{\bf{Lemma}}

\DeclareMathOperator*{\minimize}{\text{minimize}}

\usepackage{algorithm}
\usepackage{algorithmic}
\usepackage{appendix}
\usepackage[para,online,flushleft]{threeparttable}
\usepackage{multirow}
\usepackage{dcolumn}
\usepackage{booktabs}
\usepackage{makecell}

\nocopyright
\title{Towards Query-Efficient Black-Box Adversary with Zeroth-Order Natural Gradient Descent }
\author{ \Large \textbf{Pu Zhao,\textsuperscript{\rm 1} Pin-Yu Chen,\textsuperscript{\rm 2} Siyue Wang,\textsuperscript{\rm 1} Xue Lin\textsuperscript{\rm 1}}\\ % All authors must be in the same font size and format. Use \Large and \textbf to achieve this result when breaking a line
\textsuperscript{\rm 1}Northeastern University, Boston, MA 02115\\ %If you have multiple authors and multiple affiliations
\textsuperscript{\rm 2}IBM Research, Yorktown Heights, NY 10598\\
zhao.pu@husky.neu.edu, pin-yu.chen@ibm.com, wang.siy@husky.neu.edu, xue.lin@northeastern.edu % email address must be in roman text type, not monospace or sans serif
}

\begin{document}
\maketitle
\begin{abstract}
Despite the great achievements of the modern deep neural networks (DNNs), the vulnerability/robustness of state-of-the-art DNNs raises security concerns in many application domains requiring high reliability. Various adversarial attacks are proposed to sabotage the learning performance of DNN models. Among those, the black-box adversarial attack methods have received special attentions owing to their practicality and simplicity.  Black-box attacks usually prefer less queries in order to maintain stealthy and low costs. However, most of the current black-box attack methods adopt the first-order gradient descent method, which may come with certain deficiencies such as relatively slow  convergence  and  high sensitivity  to  hyper-parameter settings. In this paper, we propose a zeroth-order natural gradient descent (ZO-NGD) method to design the adversarial attacks, which incorporates the zeroth-order gradient estimation technique catering to the black-box attack scenario and the second-order natural gradient descent to achieve higher query efficiency. The empirical evaluations on image classification datasets demonstrate  that ZO-NGD can obtain significantly lower  model query complexities compared with state-of-the-art attack methods.
\end{abstract}

\section{Introduction}

Modern technologies based on %pattern recognition, 
machine learning (ML) and specifically deep learning (DL), have achieved significant breakthroughs \cite{lecun2015deep} in various applications. 
Deep neural network (DNN) serves as a fundamental component in artificial intelligence.
However, despite the outstanding performance, many recent studies demonstrate that state-of-the-art  DNNs in computer vision \cite{xie2017adversarial,7970161}, speech recognition \cite{alzantot2018did,8675229}  and  deep reinforcement learning \cite{lin2017tactics}
are vulnerable to adversarial examples \cite{goodfellow2015explaining}, which add carefully designed imperceptible distortions to legitimate inputs aiming to mislead the DNNs at test time. 
This raises concerns of the DNN robustness in many applications with high reliability and dependability requirements. % such as face recognition, autonomous driving, and malware detection \cite{mahmood2017adversarial,evtimov2017robust}. 

With the recent exploration of adversarial attacks in image classification and objection detection, the vulnerability/robustness of DNNs has attracted ever-increasing attentions and efforts in the research field known as \textit{adversarial machine learning}.  
A large amount of efforts have been devoted to: 1) designing adversarial perturbations in various ML applications \cite{goodfellow2015explaining,carlini2017towards,chen2017ead,pu2019admm,xu2018structured};  %
2) security evaluation methodologies to systematically estimate the DNN robustness \cite{biggio2014security,zhang2018efficient};  %
and 3) defense mechanisms against adversarial attacks  \cite{rota2017randomized,demontis2018yes,madry2017towards,Wang2019HRS,wang2018defensive,xu2019topology,8646578}. This work mainly investigates the first category to build  the groundwork towards developing potential defensive measures in  reliable ML.

However, most of  preliminary studies on this topic focus on the white-box setting where the target DNN model is completely available to the attacker
 \cite{goodfellow2015explaining,carlini2017towards,zhao2018admm}. 
More specifically, the adversary can compute the gradients of the output with respect to the input to identify the effect of perturbing certain input pixels, with complete knowledge about the DNN model's internal structure, parameters and configurations.
Despite the theoretical interest, it is unrealistic to adopt the white-box adversarial methods to attack practical black-box threat models \cite{pu2019fault}, where the internal model states/configurations are not revealed to the attacker (e.g., Google Cloud Vision API). Instead, the adversary can only query the model by submitting inputs and obtain the corresponding model outputs of  prediction probabilities  when generating adversarial examples.

In the black-box adversarial setting, %under the same attack threat model 
it is often the case that the less queries, the more efficient an attack becomes. Large amount of queries may be at the risk of exposing the adversary or high financial cost in the case where the query is charged per query. Notably, to date most of the white-box  \cite{goodfellow2015explaining,carlini2017towards} and black-box attacks \cite{chen2017zoo,ilyas2018blackbox,zhao2019design} are based on first-order gradient descent methods.   Different from the widely utilized first-order optimization, the application of second-order optimization \cite{martens2016second} is less explored due to the large computation overhead, although it may achieve  faster convergence rate. 
%For example, as a branch of second-order method, natural gradient descent (NGD) may enjoy higher query-efficiency as it usually requires fewer total iterations than gradient descent in many applications \cite{grosse2016kronecker}. 
The work \cite{kazuki2018second} adopts natural gradient descent (NGD) to train ResNet-50 on ImageNet in 35 epochs, demonstrating its great potentiality.

In this work, inspired by the superb convergence performance of NGD, we propose zeroth-order natural gradient descent (ZO-NGD), which incorporates the zeroth-order (ZO) method  and the second-order NGD, to generate black-box adversarial examples in a query-efficient manner.  
The  contributions of this work are summarized as follows:

+ \emph{\textbf{Design of adversary attacks with NGD:}} To the best of our knowledge, we are the first to derive the Fisher information matrix (FIM) and adopt the second-order NGD method for adversarial attacks, which is  different from other first-order-based white-box and black-box attack methods.

+ \emph{\textbf{Co-optimization of zeroth-order and second-order methods:}}  In the black-box setting, we incorporate the zeroth-order random gradient estimation to estimate the gradients which is not directly available, and leverage the second-order NGD to achieve high query-efficiency.

+ \emph{\textbf{No additional queries to obtain the FIM:}} During the queries to estimate the gradients of the loss, with our design the Fisher information is a byproduct that are extracted and evaluated without requiring additional query complexity. 

+ \emph{\textbf{Scalability to high dimensional datasets:}} 
In NGD, it is computationally infeasible to compute and invert the FIM with billions of elements on large scale datasets like ImageNet.  
To address this problem, we propose a method to avoid the computation and inverse of the FIM and thus the computation complexity is at most the same as the input images, rather than its square (the dimension of the FIM).  

\section{Related Work}\label{literature}

In adversarial ML, the black-box setting is more practical where the attacker can only query the target model by providing input images and receive the probability density output for the input. 
%Besides, under the same attack threat model, less queries is usually preferred in the black-box setting for effectiveness and stealthiness.

\subsection{Black-box Attack}

\subsubsection{Attack with gradient estimation}
In the black-box setting, as the gradients are not directly available, gradient estimation methods via zeroth-order optimization \cite{yining2018stochastic,tu2018autozoom,john2015optimal} are proposed to estimate the gradients. 
The ZOO method \cite{chen2017zoo} performs pixel-level gradient estimation first and then perform  white-box  C\&W attack  \cite{carlini2017towards}  with the estimated gradients. Despite its high success rate, it suffers from intensive computation and huge queries due to element-wise gradient estimation.

The more practical threat models are investigated in \cite{ilyas2018blackbox}. 
New attack methods based on Natural Evolutionary Strategies (NES) and Monte Carlo approximation to estimate the gradients are developed to mislead ImageNet classifiers under more restrictive threat models. %The attack methods are applied against a commercial classifier (the Google Cloud Vision API) overcoming many practical issues.
The work \cite{ilyas2018prior} further proposes to use the prior information including the time-dependent priors and data-dependent priors to enhance the query efficiency. %Then first-order gradient descent method is utilized to update the perturbation. 

Different  from the previous first-order-based methods,  the work \cite{ye2018hessianaware} exploits the second-order optimization to improve query efficiency. In general, they explore Hessian information in the parameter space while our work explores the Hessian information in the distribution space (aka information matrix). %for query efficiency. 
Particularly, our method obtains the Fisher information during the first-order information (gradients) estimation \textit{for free} while the mentioned paper needs additional queries for Hessian-based second-order optimization.

\subsubsection{Heuristic black-box attacks}

In the transfer attack \cite{nicolas2016transferability}, the attacker first trains a surrogate model with data labeled by the target model. White-box attacks are applied to attack the surrogate model and the generated examples are transferred to attack the target model. However, 
it may suffer from low attack success rate due to the low similarity between the surrogate model and the target model. 

The boundary method \cite{brendel2017decision} utilizes a conceptually simple idea  to decrease the distortion through random walks and find successful adversarial perturbations
while staying on the misclassification boundary. 
However, it suffers from  high computational complexity and  lacks algorithmic convergence guarantees.

\subsection{Second-order optimization}
First-order gradient descent methods have been extensively used in
various  ML tasks. They are easy to implement and suitable for large-scale DL.  But these methods come with well known deficiencies such as relatively-slow convergence and sensitivity to hyper-parameter settings.   
On the other hand, second-order optimization methods provide a elegant solution  by selectively re-scaling the gradient with the curvature information \cite{martens2016second}. As a kind of second-order method, NGD proves to be Fisher efficient by using the FIM instead of the Hessian matrix \cite{amari1998natural,james2014new}. But  the large overhead to compute, store and invert the FIM may limit its application. To address this, Kronecker  Factored  Approximate  Curvature (K-FAC) is proposed  to train DNNs \cite{grosse2016kronecker}. % Recently, NGD based on K-FAC has been applied to train ResNet-50 on ImageNet in 35 epochs with large mini-batch size \cite{kazuki2018second}.

\section{Problem Formulation}

\textbf{Threat Model:} In this paper, we mainly investigate black-box adversarial attacks for image classification with DNNs. Different from the white-box setting which has fully access to the DNN model and  its internal structures/parameters, 
the black-box setting constrains the information available to the adversary. The attacker can only query the  model by providing an input image and obtain the DNN output score/probability of the input.
The black-box setting is more consistent with the scenario of ``machine-learning-deployed-as-a-service'' like Google Cloud Vision API. 

In the following, we first provide a general problem formulation for adversarial attack which can be adopted to either white-box or black-box settings. 
Then,  an efficient solution is proposed for the black-box setting. We highlight that this method can be easily adopted to the white-box setting by using the exact gradients to achieve higher query efficiency.

\textbf{Attack Model:} Given a legitimate image $\mathbf x \in \mathbb R^d$ with its correct class label $t$, the objective is to design an optimal adversarial perturbation $\bm{\delta} \in \mathbb R^d$ so that the perturbed example $(\mathbf x + \boldsymbol{\delta})$ can lead to a misclassification by the DNN model trained on legitimate images. The DNN model would misclassify the adversarial example to another class $t' \neq t$.  $\boldsymbol \delta$ can be obtained by solving the following problem,
{\small \begin{align}\label{eq: prob}
    \displaystyle \minimize_{\boldsymbol \delta \in \mathbb S}  f(\mathbf x + \boldsymbol \delta, t),
\end{align}}%
where
%\begin{equation}
$\mathbb S=\{ \bm \delta |  (\mathbf x + \boldsymbol \delta) \in [0,1]^d, ~ \| \boldsymbol \delta \|_\infty \leq \epsilon \} $
%\end{equation}
and $f(\mathbf x+ \bm \delta, t)$ denotes an attack loss incurred by misclassifying $(\mathbf{x} + \boldsymbol{\delta})$ to another class $t'$.
 $\| \cdot \|_\infty$ denotes  the $\ell_\infty$ norm.
In problem \eqref{eq: prob}, the  constraints on $\mathbb S$ ensure that the perturbed noise $\boldsymbol{\delta}$ at each pixel (normalized to $[0,1]$) is imperceptible up to a predefined $\epsilon$-tolerant threshold.  

Motivated by \cite{carlini2017towards}, 
 the loss function $f(\bm x, t) $ is expressed as
{\small\begin{align}\label{eq: fx}
f(\mathbf x + \boldsymbol \delta, t) \ =\ \max \{ & \log p(t| \mathbf x + \boldsymbol \delta )
\nonumber  \\
& - \max_{i \neq t} \{ \log p(i| \mathbf x + \boldsymbol \delta)\} , - \kappa \},
\end{align}}%
where $p(i| \mathbf x)$ denotes the model's prediction score or probability of the $i$-th class for the input $\bm x$,  and $\kappa$ is a confidence parameter usually set to zero.  
Basically,  $f(\mathbf x + \boldsymbol \delta, t)$ achieves its minimum value 0 if  $p(t| \mathbf x + \boldsymbol \delta )$  is smaller than $\max_{i \neq t} \log p(i| \mathbf x + \boldsymbol \delta)$, indicating there is a label with higher probability than the correct label $t$ and thus a misclassification is achieved by adding the perturbation $\bm \delta$ to $\bm x$. In this paper, we mainly investigate the untargeted attack which does not specify the target misclassified label. The targeted attack can be easily implemented following nearly the same problem formulation and loss function with slight modifications \cite{carlini2017towards,ilyas2018prior}. We focus on the general formulation here and omit the targeted attack formulation. 

Note that in Eq.~\eqref{eq: fx}, we use the log probability $\log p(i| \mathbf x)$  instead of
$p(i|\mathbf{x})$ 
because the output probability distribution   tends to have one dominating class.  
The log operator is used to reduce the effect of the  dominating class while it still preserves the probability order of all classes.

As most of the white-box attack methods rely on gradient descent methods, the unavailability of the gradients in black-box settings will limit their application. Gradient estimation methods (known as zeroth-order optimization) are applied to perform the normal projected first-order  gradient descent process \cite{ilyas2018blackbox,liu2018signsgd} as follows,
 {\small\begin{equation}
{{\bm{\delta }}_{k + 1}} = {\prod _{\mathbb  S}}\left( {{{\bm{\delta }}_k} - \lambda \hat \nabla f({{\bm{\delta }}_k})} \right),
\end{equation}}%
where  $\lambda$ is the learning rate and the $\prod_{\mathbb  S}(\cdot)$ performs the projection onto the feasible set $\mathbb S$. %Basically it follows the gradient descent.

In the black-box setting,  it is often the case that the query number is limited or high query efficiency is required by the adversary. 
The zeroth-order method  tries to extract gradient information of the objective function and the first-order method is applied to minimize the loss due to its wide application in ML. However, the second-order information of the queries is not fully exploited.
In this paper, we aim to take advantages of the model’s second-order information and propose a novel method named ZO-NGD optimization.

\section{Zeroth-order Nature Gradient Descent}

The proposed method is based on  NGD \cite{james2014new} and ZO optimization  \cite{john2015optimal}.
In the applications of optimizing probabilistic
models, NGD uses the natural gradient  by multiplying the gradient with the FIM  %instead of the standard gradient 
to update the parameters. 
NGD seems to be a potentially attractive alternative method as it requires  fewer  total iterations than gradient descent \cite{ollivier2015riemannian,martens2015optimizing,grosse2015scaling}. 

Motivated from the perspective of information geometry, NGD  defines the steepest descent/direction in the realizable  distribution space instead of the  parameter space. 
The distance in the distribution space is measured with a special “Riemannian metric” \cite{amari2007methods}, which is different from the standard Euclidean distance metric in the parameter space. 
This Riemannian metric does not rely on the parameters like the Euclidean metric, but depends on the distributions themselves. Thus it is invariant to any smooth or invertible reparameterization of the model.
More details are discussed in the Geometric Interpretation Section.

Next we will introduce the FIM and the implementation details to perform NGD. Basically, 
the proposed framework first queries the model to estimate the gradients and Fisher information. Then after the damping and inverting processes, natural gradient is obtained to update the perturbation. Algorithm \ref{alg:Framwork} shows the pseudo code of the ZO-NGD.

\begin{algorithm}[tb] 
\caption{ Framework of ZO-NGD.} \label{alg:Framwork} 
\begin{algorithmic}[1] 
\REQUIRE ~~\\ 
The legitimate image $\bm x$; the correct label $t$; the model to be queried; the learning rate $\lambda$; the sampling step size $\mu$;
\ENSURE ~~\\ 
Adversarial perturbation $\bm \delta$; 
\STATE initialize $\bm \delta_0$ with all zeros;
\FOR {$k=0,...,K$}
\STATE Query the model with $\bm \delta_k$ and obtain the probability $p(t|\bm x, \bm \delta_k):=p(t|\bm x+ \bm \delta_k)$;
\FOR{$j=1,...,R$}
\STATE Generate a random direction vector $ \mathbf u_j$  drawn from a uniform distribution over the surface of a unit sphere;
\STATE Query the model with $\bm x+ \bm \delta_k + \mu \mathbf u_j$ and obtain  $p(t|\bm x, \bm \delta_k + \mu \mathbf u_j)$;
\ENDFOR
\STATE Estimate the gradients of the loss function $\hat \nabla f(\boldsymbol{\delta}_k)$ according to Eq. \eqref{eq: grad_rand_ave_loss};
\STATE Estimate the gradients of the log-likelihood function $\hat \nabla \log p(t\left| {{\bf{x}},{\bf{\delta }_k}} \right.)$  according to Eq. \eqref{eq: grad_rand_ave_log-like};
\STATE  Compute the FIM $ \bf F$ according to Eq. \eqref{eq: estimate_FIM} and  perform the nature gradient update  as shown in Eq. \eqref{eq: ZO_nature_grad_update}. 
\ENDFOR
\end{algorithmic} 
\end{algorithm}

\subsection{Fisher Information Matrix and Natural Gradient}
We introduce and derive the FIM in this section. In general, finding an adversarial example can be formulated as a training problem. In the idealized setting, input vectors $\bm x$ are drawn independently from a distribution $Q_{\bm x}$ with density function $q(\bm x)$, and the corresponding output $ t$ is drawn from a conditional target distribution ${Q_{ t\left| {\bf{x}} \right.}}$ with density function $q( t\left| {\bm x} \right.)$. The target joint distribution is ${Q_{t,{\bf{x}}}}$ with the  density of  $q(t, \bm x) = q(t| \bm x)q(\bm x)$. By finding an adversarial perturbation $\bm \delta$, we obtain the learned distribution ${P_{t,{\bf{x}}}(\bm \delta)}$, whose density is $p(t, \bm x| \bm \delta) =p(t|\bm x + \bm \delta)q(\bm x) := p(t|\bm x, \bm \delta)q(\bm x)$.

In statistics, the score function  \cite{cox1979theoretical} indicates how sensitive a likelihood function $p(t, \bm x| \bm \delta)$ is to its parameters $\bm \delta$. Explicitly, the score function for $\bm \delta$ is the gradient of the log-likelihood with respect to $\bm \delta$ as% specified
below,
{\small\begin{equation}
s({\bm{\delta }}) = \nabla \log p(t, \bm x| \bm \delta).
\end{equation}}%

\begin{lemma}  \label{score-expectation}
The expected value of the score function with respect to $\bm \delta$ is zero.
\end{lemma}
The proof is shown in the appendix. We can define an uncertainty measure around the expected value (i.e., the covariance of the score function) as follows,
 {\small\begin{equation}
\mathbb E\left[ {\left( {s({\bm{\delta }}) - \mathbb E[s({\bm{\delta }})] } \right){{\left( {s({\bm{\delta }}) - \mathbb E[s({\bm{\delta }})]} \right)}^{\rm{T}}}} \right].
\end{equation}}%
The covariance of the score function above is the definition of the Fisher information. It is in the form of a matrix and the FIM can be written as 
{\small\begin{equation} 
\mathbf{F}=\mathbb E_{\bm x\in Q_x}\left[ \mathbb E_{\tilde{t}\sim p\left( \cdot \left|\bm  x, \bm \delta \right. \right)}\left[ \nabla \log p\left( \tilde{t}, \bm  x\left|\bm \delta \right. \right) \nabla \log p\left( \tilde{t}, \bm x \left| \bm \delta \right. \right) ^T \right] \right].
\end{equation}}%
Note that this expression involves the losses on all possible values of the classes $ \tilde t$, not only the actual label for each data sample. As $\bm \delta$ only corresponds to a single input $\bm x$, the training set only contains one data sample.
Besides, since $p(\tilde t, \bm x| \bm \delta) =p(\tilde t|\bm x + \bm \delta)q(x) = p(\tilde t|\bm x, \bm \delta)q(\bm x)$ and $q(\bm x)$ does not depend on $\bm \delta$, we have
{\small\begin{align}
\nabla \log p(\tilde t, \bm x| \bm \delta)  &= \nabla \log  p(\tilde t|\bm x, \bm \delta) + \nabla \log  q(\bm x) \nonumber \\ 
&=\nabla \log  p(\tilde t|\bm x, \bm \delta).
\end{align}}%
Then the FIM can be transformed to
{\small\begin{equation}
\mathbf{F} = \mathbb E_{\tilde{t}\sim p\left( \cdot \left| \bm x, \bm \delta \right. \right)} \left[ {\nabla \log p(\tilde{t}|\bm x, \bm \delta)\nabla \log p{{(\tilde t|\bm x, \bm \delta)}^{\rm{T}}}} \right].
\end{equation}}%
%The expectation is taken over all possible values of the labels $ \tilde t$.  
The exact expectation with $T$ categories is expressed as, 
{\small\begin{equation} \label{equ: precise_Fish}
\mathbf{F}=\sum_{\tilde{t}=1}^T{p\left( \tilde{t}\left| \bm x,\bm \delta \right. \right) \nabla \log p\left( \tilde{t}\left| \bm x, \bm \delta \right. \right) \nabla \log p\left( \tilde{t}\left| \bm  x, \bm \delta \right. \right) ^T}
\end{equation}}%

The usual definition of the natural gradient is % which appears in the literature is
{\small\begin{equation}
\tilde \nabla f({\bm{\delta }})={{\bf{F}}^{ - 1}} \nabla f({\bm{\delta }}),
\end{equation}}%
and the NGD minimizes the loss function through 
{\small\begin{equation}
{{\bm{\delta }}_{k + 1}} = {{\bm{\delta }}_k} - \lambda \tilde \nabla f({{\bm{\delta }}_k}).
\end{equation}}%

\subsection{Outer Product and Monte Carlo Approximation}

The FIM involves an expectation over all possible classes ${\tilde{t}\sim p\left( \cdot \left| x, \delta \right. \right)}$ drawn from the   probability distribution output. 
 In the case with large number of classes, it is impractical to compute the exact FIM due to the  intensive computation. To address the high computation overhead, %in the case of large number of $\tilde t$, 
in general there are two methods to approximate the FIM, the outer product approximation and the Monte Carlo approximation.

\subsubsection{Outer Product Approximation}
The outer product approximation of the FIM \cite{razvan2013natural,ollivier2015riemannian} only uses the actual label $t$ %in the dataset  
to avoid the expectation over all possible labels ${\tilde{t}\sim p\left( \cdot \left| x, \delta \right. \right)}$, as below,
{\small\begin{equation} \label{equ: out_product}
\mathbf{F}_{OP}=\nabla \log p\left( {t}\left| \bm x,\bm \delta \right. \right) \nabla \log p\left( {t}\left|\bm  x,\bm \delta \right. \right) ^T.
\end{equation}}%
Thus  a rank-one matrix can be obtained directly.% by the outer product of the gradient. % provided by usual backpropagation. 

\subsubsection{Monte Carlo Approximation}
Monte Carlo (MC) approximation \cite{ollivier2015riemannian} replaces the expectation over $\tilde t$ with $n_{MC}$ samples,
{\small\begin{equation}
\mathbf{F}_{MC}=\frac{1}{n_{MC}}\sum_{i=1}^{n_{MC}}{\nabla \log p\left( \tilde{t}_i\left| \bm  x,\bm \delta \right. \right) \nabla \log p\left( \tilde{t}_i\left|\bm  x,\bm \delta \right. \right) ^T}.
\end{equation}}%
where each  $\tilde t_i$ is drawn from the distribution $ {p\left( \cdot \left| x, \delta \right. \right)}$. The MC natural gradient  works well in practice with $n_{MC}= 1 $.

For higher query efficiency, we adopt the outer product approximation as it does not require additional queries .

\subsection{Gaussian Smoothing and Gradient Estimation}
To compute the FIM and perform NGD,
we need to obtain the gradients of the loss function $\nabla f(\bm \delta)$ and the gradients of the log-likelihood $\nabla p{(t|\bm x, \bm \delta)}$,  which are not directly available in the black-box setting.  

To address this difficulty, we first introduce the Gaussian approximation of $f(\bm x)$  \cite{nesterov2017random},
{\small\begin{equation}
f_{\mu}\left( \bm x \right) =\frac{1}{\left( 2\pi \right) ^{d/2}}\int_{R^d}{f\left( \bm x+\mu \bm u \right) \exp \left( -\frac{1}{2}\lVert \bm u \rVert ^2 \right) d \bm u},
\end{equation}}%
where  ${\lVert \cdot \rVert}$ is the Frobenius norm, $\mu> 0$ is a smoothing parameter and $\bm u$ is a random vector distributed uniformly over the \textcolor{black}{surface of} a unit sphere, i.e., $\bm u\sim N\left( \text{0,} \mathbf{I}_d \right) $. %\textcolor{blue}{PY: Note that Gauss rv is NOT distributed on unit sphere. Some inconsistecy here.}  \textcolor{red}{ It should be the surface of the unit sphere. Then it's Gauss.}
Its gradient can be written as 
{\small\begin{align} \label{eq: gradient_Gau_smooth} 
\nabla f_{\mu}\left(\bm  x \right) & =\frac{1}{M}\int_{R^d}{\frac{f\left( \bm x+\mu \bm u \right) -f\left( \bm x \right)}{\mu}\bm u\exp \left( -\frac{1}{2}\lVert \bm u \rVert ^2 \right) d \bm u}
  \nonumber \\
& =E_{\bm u}\left( \frac{f\left(\bm x+\mu \bm u \right) -f\left( \bm x \right)}{\mu}\bm u \right),
\end{align}}%
where $E_{\bm u}$ is the Gaussian smoothing function.
%\textcolor{blue}{ $x$ and $u$ should be boldface }
Thus, based on Eq. \eqref{eq: gradient_Gau_smooth}, we apply the  zeroth-order  random gradient estimation to estimate the gradients by
 {\small\begin{equation}\label{eq: grad_rand_ave_loss}
\hat \nabla f(\bm{\delta}) = \frac{1}{R} \sum_{j=1}^R 
 \frac{f ( \boldsymbol{\delta} + \mu \mathbf u_{j}, t) - f ( \boldsymbol{\delta}, t ) }{\mu}   \mathbf u_{j}, 
\end{equation}}%
and 
 {\small\begin{align}
 \label{eq: grad_rand_ave_log-like}
\hat \nabla \log p(t\left| {{\bf{x}},{\bm{\delta }}} \right.) = \frac{1}{R \mu}\sum\limits_{j = 1}^R [\log p(t\left| {{\bf{x}},{\bm{\delta }}} \right. + \mu {{\bf{u }}_j})
\nonumber \\
- \log p(t\left| {{\bf{x}},{\bm{\delta }}} \right.)] {{\bf{u }}_j},
\end{align}}%
where
$R$ is the number of random direction vectors and $\{ \mathbf u_j \}$ denote independent and identically distributed (i.i.d.) random direction vectors following Gaussian distribution. 
 
We note that in each gradient estimation step, by querying the model $R+1$ times, we can \textit{simultaneously} obtain both the $\hat \nabla f(\bm \delta)$ and $\hat \nabla \log p(t\left| {{\bf{x}},{\bm{\delta }}} \right.)$ as demonstrated in Algorithm 
\ref{alg:Framwork}. Different from the zeroth-order gradient descent which only estimates the gradients of the loss function $\hat \nabla f(\bm \delta)$ (such as \citeauthor{chen2017zoo} and \citeauthor{ilyas2018blackbox}), ZO-NGD obtains $\hat \nabla \log p(t\left| {{\bf{x}},{\bm{\delta }}} \right.)$ and computes the FIM from the same query outputs without incurring additional query complexity. This is one major difference between ZO-NGD and other zeroth-order methods. 
Thus, higher query-efficiency can be achieved by leveraging the FIM and second-order optimization.

\subsection{Damping for Fisher Information Matrix}
The inverse of the FIM is required for natural gradient. However,
the eigenvalue distribution of the FIM  is known to have an extremely long tail  \cite{karakida2018universal}, where most of the eigenvalues are close to zero. This in turn causes the eigenvalues of the inverse FIM to be extremely large,   leading to the  unstable training. To mitigate this problem, damping technique is used to add a positive value to the diagonal of the FIM to stabilize the training as shown below,
{\small\begin{equation}\label{eq: estimate_FIM}
\mathbf{\hat{F}}=\hat \nabla \log p\left( t\left| \bm x,\bm \delta \right. \right) \hat \nabla \log p\left( t\left| \bm x,\bm \delta \right. \right) ^T+\gamma \mathbf{I},
\end{equation}}%
where $\gamma$ is a constant. As the damping limits the maximum eigenvalue of the inverse FIM, we can restrict the norm of the gradients. This prevents ZO-NGD from moving too far in flat directions.

With the obtained FIM, the perturbation  update is 
{\small\begin{equation}\label{eq: ZO_nature_grad_update}
{{\bm{\delta }}_{k + 1}} = {\prod _{\mathbb  S}}\left({{\bm \delta }_k} - \lambda \mathbf{\hat{F}}^{ - 1}   \hat \nabla f({{\bm{\delta }}_k}) \right).
\end{equation}}%
ZO-NGD tries to extract the Fisher information to perform second-order optimization  for faster convergence rate and better query efficiency.

\subsection{Scalability to High Dimensional Datasets }
Note that the FIM has a dimension of $d^2$ where $d$ is the dimension of the input image.  On ImageNet dataset which typically contains images with about $270,000$ pixels ($\bm x \in \mathbb R ^{299\times 299 \times 3} $), the FIM would have billions of elements and thus it is quite difficult to compute or store the FIM, not to mention its inverse.
In the application of training DNN models, the Kronecker Factored Approximate Curvature (K-FAC) method \cite{martens2015optimizing,kazuki2018second} is adopted to deal with the difficulty of high dimensions of the DNN model. However, K-FAC methods may not be suitable in the application of finding adversarial examples as the assumption of uncorrelated channels is not valid and thus we can not apply the block diagonalization  method for the FIM.
Instead, we propose another method to compute $\mathbf{\hat{F}}^{ - 1} $ of high dimensions as follows. 
First we have  
{\small\begin{equation}
\hat \nabla \log p\left( {t}\left| \bm x, \bm \delta \right. \right) =c\frac{\hat \nabla \log p\left( {t}\left| \bm x, \bm \delta \right. \right)}{\lVert \hat \nabla \log p\left( {t}\left| \bm x, \bm \delta \right. \right) \rVert} = c \frac{ \hat s(\bm \delta)}{\lVert \hat s(\bm \delta)  \rVert},
\end{equation}}%
where $c=\lVert \hat s(\bm \delta)  \rVert$. The inverse matrix $\mathbf{\hat{F}}^{-1}$ can be represented as,
{\small\begin{equation}
\mathbf{\hat{F}}^{-1}=\frac{\left( \left( c^2+\gamma \right) ^{-1}-\gamma ^{-1} \right)}{c  ^2}\hat{s}\left( \bm \delta \right) \hat{s}\left( \bm \delta \right) ^T+\gamma ^{-1} \mathbf{I}. 
\end{equation}}%
This can be  verified simply by checking their multiplication and we omit the proof here. Then the gradient update $  \varDelta \bm \delta  =  \lambda  \mathbf{\hat{F}}^{-1}  \nabla f( \bm \delta) $ in Eq. \eqref{eq: ZO_nature_grad_update} is 
{\small\begin{equation} \label{eq: delta_update}
\begin{aligned}
\varDelta \bm \delta & =\lambda \left[ \frac{\left( \left( c^2+\gamma \right) ^{-1}-\gamma ^{-1} \right)}{c ^2}\hat{s}\left( \bm  \delta \right) \hat{s}\left( \bm  \delta \right) ^T+\gamma ^{-1} \mathbf{I} \right] \hat \nabla f\left(\bm  \delta \right)
\\
\,\,  & =\lambda \frac{\left( \left( c^2+\gamma \right) ^{-1}-\gamma ^{-1} \right)}{c ^2}\hat{s}\left( \bm \delta \right) \left[ \hat{s}\left(\bm  \delta \right) ^T \hat \nabla f\left( \bm \delta \right) \right] 
\\ & \ \ \ \ +\lambda \gamma ^{-1}\hat \nabla f\left( \bm \delta \right).
\end{aligned}
\end{equation}}%
During the computation of  $\varDelta \bm \delta = \lambda  \mathbf{\hat{F}}^{-1}  \nabla f( \bm \delta)$, we compute $\hat{s}\left( \bm \delta \right) ^T \hat \nabla f\left( \bm \delta \right)$ first in Eq. \eqref{eq: delta_update} and obtain a scalar, then $\varDelta \bm \delta$ is simply  the sum of two vectors. Although $\mathbf{\hat{F}}$ and its inverse might have  billions of elements, we avoid directly computing them and the dimension of the internal computation is at most the same level as the dimension $d$ of the images, rather than its square $d^2$. Thus, the ZO-NGD method can be applied on datasets with high dimensional images.

\subsection{Geometric Interpretation} \label{geometric_interpretation}
We provide a geometric interpretation for the natural gradient here.
The negative gradient $- \nabla f (\bm \delta)$ can be interpreted as the steepest descent direction  in the sense that it yields the most reduction in  $f$ per unit of change of $\bm \delta$, where the change is measured by the standard Euclidean norm $\left\|  \cdot  \right\|$ \cite{james2014new}, as shown below,
{\small\begin{equation}
\frac{{ - \nabla f(\bm \delta)}}{{\left\| {\nabla f(\bm \delta)} \right\|}} = \mathop {\lim }\limits_{\epsilon  \to 0} \frac{1}{\epsilon }\arg \mathop {\min }\limits_{\left\| \alpha  \right\| \le \epsilon } f(\bm \delta  + \bm \alpha ).
\end{equation}}%
By  following the $- \nabla f(\bm \delta)$ direction, we can obtain the change of $\bm \delta$ within a certain $\epsilon$-neighbourhood to minimize the loss function. 

\begin{lemma}  \label{NG-distribution}
The negative natural gradient is the steepest descent direction in the distribution space.
\end{lemma}

We provide the proof of Lemma \ref{NG-distribution} in the appendix\footnotemark[1]. In the parameter space, the negative gradient is the steepest descent direction to minimize the loss function. By contrast, in the distribution space where the distance is measured by KL divergence, the steepest descent direction  is  the negative  natural gradient. Thus, the direction in distribution space defined  by the natural gradient will be invariant to the choice of parameterization \cite{pascanu2013revisiting}, i.e., it will not be affected by how the model is parametrized, but only depends on the distribution induced by the parameters.

\section{Experimental Results}
In this section, we present the experimental results of the ZO-NGD method. 
We compare ZO-NGD with 
various attack methods on three image classification datasets, MNIST \cite{Lecun1998gradient}, CIFAR-10 \cite{Krizhevsky2009learning} and ImageNet \cite{deng2009imagenet}.

We train two networks for MNIST and CIFAR-10 datasets, respectively.
The model for MNIST achieves 99.6\% accuracy  with four convolutional layers, two max pooling layers, two fully connected layers and a softmax layer. For CIFAR-10, we adopt the same model architecture as MNIST, achieving 80\% accuracy. For ImageNet, a pre-trained Inception v3 network \cite{Szegedy2016RethinkingTI} is utilized  instead of training our own model, attaining 96\% top-5 accuracy. 
All experiments are performed on machines with 
NVIDIA GTX 1080 TI GPUs.

%\subsubsection{General Results}
\subsection{Evaluation of White-box Attack}
We first check  the white-box setting, where we compare the proposed NGD with PGD from adversarial training. PGD is a typical first-order method while NGD utilizes the second-order FIM. The query here is defined as one forward pass and one subsequent backpropagation as we need to obtain the gradients through backpropagation. We report the average number of queries over 500 images for successful adversaries on each dataset. On MNIST, NGD requires 2.12 queries while PGD needs 4.88 queries with $\epsilon=0.2$.  On CIFAR-10, NGD requires 2.06 queries while PGD needs 4.21 queries with $\epsilon=0.1$.  On ImageNet, NGD requires 2.20 queries while PGD needs 5.62 queries with $\epsilon=0.05$. We can see that NGD achieves higher query efficiency by incorporating FIM.

\subsection{Evaluation on MNIST and CIFAR-10}

In the evaluation on MNIST and CIFAR-10, we select 2000 correctly classified images from MNIST and CIFAR-10 test datasets, respectively, and perform black-box attacks for these images. We compare the ZO-NGD method with the transfer attack \cite{nicolas2016transferability},  ZOO black-box attack \cite{chen2017zoo}, and the natural-evolution-strategy-based projected gradient descent method (NES-PGD) \cite{ilyas2018blackbox}.  For the transfer attack \cite{nicolas2016transferability}, we apply C\&W attack \cite{carlini2017towards}  to the surrogate model. 
The implementations of ZOO and NES-PGD are based on the GitHub code released by the authors\footnote{\label{note1}%https://github.com/carlini/nn\_robust\_attacks
The code and appendix are available at \url{https://github.com/LinLabNEU/ZO_NGD_blackbox}.
}. 
For the attack methods, the pixel values of all images are normalized to the range of $[0,1]$.
In the proposed ZO-NGD method,  the sampling number $R$ in the random gradient estimation as defined in Eq. \eqref{eq: grad_rand_ave_loss} and \eqref{eq: grad_rand_ave_log-like} is set to 40.
 $\epsilon$ is set to 0.4  for MNIST and  0.2 for CIFAR-10  or ImageNet.
  In Eq. \eqref{eq: grad_rand_ave_loss} and \eqref{eq: grad_rand_ave_log-like}, we set $\mu=1$   for three datasets. $\gamma$ is set to 0.01.

  \begin{table} [tb]
 \centering
  \caption{Performance evaluation of black-box adversarial attacks on MNIST and CIFAR-10.}
  \label{table_MNIST}
  \scalebox{0.95}[0.95]{
   \begin{threeparttable}
\begin{tabular}{c|c|c|c|c}
    \hline
\toprule[1pt]
dataset & Attack method & 
\makecell{success\\rate} &  \makecell{average \\ queries }&  \makecell{reduction\\ rate}    \\
\midrule[1pt]
\multirow{4}{*}{MNIST} & \makecell{Transfer attack} %\cite{nicolas2016transferability} 
 & 82\% & - & - \\
%\cline{2-4} 
&ZOO attack %\cite{chen2017zoo} 
& 100\% & 8,300 & 0\% \\ 
%\cline{2-4} 
&NES-PGD %\cite{ilyas2018blackbox}  
& 98.2\%  & 1,243  & 85\% \\ 
%\cline{2-4} 
& ZO-NGD & 98.7\% & 523 &93.7\% \\ 
\midrule[1pt]
\multirow{4}{*}{CIFAR}  & \makecell{Transfer attack} %\cite{nicolas2016transferability} 
& 85\%  & -  & - \\
%\cline{2-5} 
 & ZOO attack %\cite{chen2017zoo} 
 & 99.8 \% &  6,500  & 0\% \\ 
%\cline{2-5} 
& NES-PGD %\cite{ilyas2018blackbox} 
& 98.9\%& 417 & 93.6\% \\ 
%\cline{2-5} 
 & ZO-NGD &  99.2 \%& 131 & 98\%\\ 
\bottomrule[1pt]
  \end{tabular}
\end{threeparttable}} 
\end{table}

  The experimental results are summarized in Table  \ref{table_MNIST}. %and Table \ref{table_CIFAR}. 
  We show the success rate and the average queries over successful adversarial examples for the black-box attack methods on MNIST and CIFAR-10 datasets. As shown in Table \ref{table_MNIST}, % and  \ref{table_CIFAR}, 
  the transfer attack does not achieve high success rate due to the difference between the surrogate model and the original target model. %Besides, it requires many queries to obtain the target model's outputs and train the surrogate model with these outputs. 
  The ZOO attack method can achieve high success rate at the cost of excessive query complexity since it performs gradient estimation for each pixel of the input image.  We can observe that the ZO-NGD method requires significantly less queries than the NES-PGD method.  NES-PGD uses natural evolutionary strategies for gradient estimation and then perform first-order gradient descent to obtain the adversarial perturbations.  Compared with NES-PGD, the proposed ZO-NGD not only estimates the first-order gradients of the loss function, but also  tries to obtain the second-order Fisher information from the queries without incurring additional query complexities, leading to higher query-efficiency. From Table \ref{table_MNIST}, % and \ref{table_CIFAR}, 
  we can observe that  the ZO-NGD method attains the smallest number of queries to successfully obtain the adversarial examples in the black-box setting. Benchmarking on the ZOO method, the query reduction ratio of ZO-NGD can be as high as 93.7\% on MNIST and 98\% on CIFAR-10. 
  
\subsection{Evaluation on ImageNet}

We perform black-box adversarial attacks on ImageNet %and demonstrate the performance in this section.
where 1000 correctly classified images are randomly selected. On ImageNet, we compare the proposed ZO-NGD with the ZOO attack, NES-PGD method and the bandit attack with time and data-dependent priors (named as Bandits[TD]) \cite{ilyas2018prior}. The transfer attack is not performed since it is not easy to train a surrogate model on ImageNet. The Bandits[TD] method makes use of the prior information for the gradients estimation, including the time-dependent priors which  explores the heavily correlated successive gradients, and the data-dependent priors which  exploits the spatially local similarity exhibited in images. After gradient estimation with the priors or bandits information, first-order gradient descent method is applied. 

\begin{table} [tb]
 \centering
  \caption{Performance evaluation of black-box adversarial attacks on ImageNet.}
  \label{table_ImageNet}
  \scalebox{0.95}[0.95]{
   \begin{threeparttable}
\begin{tabular}{c|c|c|c}
\toprule[1pt]
 Attack method & \makecell{ success \\ rate } &  \makecell{average \\ queries}   &  \makecell{reduction \\ ratio} \\ 
\midrule[1pt]
 ZOO attack
 & 98.9\%  & 16,800 & 0\%\\ 
\hline
 NES-PGD 
 & 94.6\% & 1,325 &  92.1\%\\ 
\hline
Bandits[TD] 
& 96.1\%  &  791  & 95.3\%   \\ 
\hline
 ZO-NGD & 97\%  & 582 & 96.5\% \\ 
\bottomrule[1pt]
 \end{tabular}
\end{threeparttable}}  
\end{table} 

\begin{figure}[tb]    
 \centering
\includegraphics[width=.95\columnwidth]{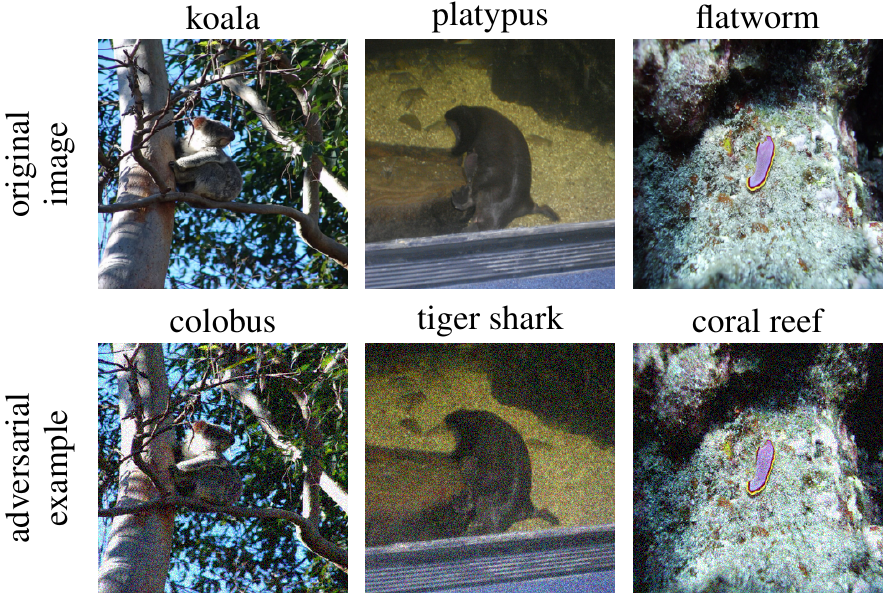}
\caption{\footnotesize{ The legitimate images and their adversarial examples generated by ZO-NGD.
}} \label{fig: examples}
\end{figure}

We present the performance evaluation on ImageNet in Table  \ref{table_ImageNet}. The success rate and the average queries over successful attacks for various black-box attack methods are reported.
Table \ref{table_ImageNet} shows the ZOO attack method can achieve high success rate with high query complexity due to its element-wise gradient estimation.  
We can have a similar observation that the ZO-NGD method only requires a much smaller number of queries than the NES-PGD method due to the faster convergence rate of second-order optimization by  exploring the Fisher information.  We also find that the ZO-NGD method also outperforms the Bandits[TD] method in terms of query efficiency. 
The Bandits[TD] method enhances the query efficiency of gradient estimations through the incorporation of priors information for the gradients, but its attack methodology is still based on the first-order gradient descent method.
As observed from Table \ref{table_ImageNet},   the ZO-NGD method achieves the highest query-efficiency for successful adversarial attacks in the black-box setting. It can obtain 96.5\%   query reduction ratio on ImageNet when compared with the ZOO method.
In Figure \ref{fig: examples}, we show some legitimate images on ImageNet and their corresponding adversarial examples obtained by ZO-NGD. We can observe that the adversarial perturbations are imperceptible. More examples on MNIST and CIFAR-10 are shown in the appendix.

% We note that the ASR of ZOO is higher than ZO-NGD. ZOO method estimates the gradients pixel by pixel at the cost of huge queries. The relatively small estimation error leads to high ASR. Besides, in the ZO-NGD experiments, we set up different maximum query number for ZOO and ZO-NGD. Both algorithms stop if it reaches the maximum value. To make a fair comparison, we further investigate a query-limited setting here, where the maximum query number (MQ) for ZOO and ZO-NGD are set to the same value.  
% We report the ASR for 500 images on ImageNet in this setting. By setting MQ to 10000, ZO-NGD achieves an ASR of 97.4\% while ZOO achieves 42.2\% ASR.  By setting MQ to 20000, ZO-NGD achieves an ASR of 98.4\% while ZOO achieves 95.6\% ASR.  By setting MQ to 30000, ZO-NGD achieves an ASR of 98.8\% while ZOO achieves 98.6\% ASR. So in the query limited setting, it is harder for ZOO to achieve high ASR if the query limit is tight. If given large enough queries, the two methods have similar performance on ASR. However, ZO-NGD requires much less average queries than ZOO as shown in the paper.  

\subsection{Ablation study}
 In this ablation study, we perform sensitivity analysis on the proposed ZO-NGD method based variations in 
 model architectures and different parameter settings. Below we summarize the conclusion and findings from this ablation study and report their details in the appendix. (1) Tested on VGG16 and ResNet and varying the parameters $\mu$ and $\epsilon$ in ZO-NGD, the results
 demonstrate the consistent superior performance of ZO-NGD by leveraging the second-order optimization. (2) We  inspect the approximation techniques used in ZO-NGD including damping and outer product method. The results show that there is a wide range of proper $\gamma$ values such that damping can work effectively to reduce the loss, and the outer product is a reasonable approximation based on the empirical evidence. We also note that the ASR of ZOO is higher than ZO-NGD. We provide a discussion about the ASR v.s. query number in the appendix.

\subsubsection{Query Number Distribution} 
\begin{figure}[h]
\centering
  \includegraphics[width=0.7\linewidth]{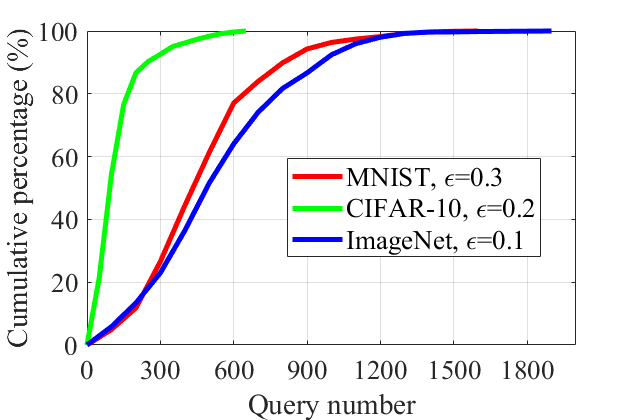}
  %\vspace{-1mm}
  \caption{\footnotesize{CDF of query number on three datasets using ZO-NGD.}} 
  \label{fig: query_distribution}
\end{figure}

Figure \ref{fig: query_distribution} shows the cumulative distribution (CDF) of the query number for 1000 images on three datasets, validating ZO-NGD's query efficiency.

\subsection{Transferability}
The transferability of adversarial examples is an interesting and valuable metric to measure the performance. To show the transferability, we use 500 targeted adversarial examples  generated by ZO-NGD  on ImageNet with  $\epsilon= 0.1$ on Inception to attack ResNet and VGG16 model. It achieves 94.4\% and 95.6\% ASR, respectively, demonstrating high transferability of our method.  Our transferred ASR is also higher than NES-PGD (92.1\% and 92.9\% ASR).

\section{Conclusion}
In this paper, we propose a novel ZO-NGD to achieve high query-efficiency in black-box adversarial attacks. It incorporates the ZO random gradient estimation and the second-order FIM for NGD. The performance evaluation on three image classification datasets demonstrate the effectiveness of the proposed method in terms of fast convergence and improved query efficiency over state-of-the-art methods. 

\section{Acknowledgments}
This work is partly supported by the National Science Foundation CNS-1932351, and is also based upon work partially supported by the Department of Energy National Energy Technology Laboratory under Award Number DE-OE0000911.

\section{Disclaimer}
This report was prepared as an account of work sponsored by an agency of the United States Government. Neither the United States Government nor any agency thereof, nor any of their employees, makes any warranty, express or implied, or assumes any legal liability or responsibility for the accuracy, completeness, or usefulness of any information, apparatus, product, or process disclosed, or represents that its use would not infringe privately owned rights. Reference herein to any specific commercial product, process or service by trade name, trademark, manufacturer, or otherwise does not necessarily constitute or imply its endorsement, recommendation, or favoring by the United States Government or any agency thereof. The views and opinions of authors expressed herein do not necessarily state or reflect those of the United States Government or any agency thereof.

{\small
\bibliographystyle{aaai}
\bibliography{egbib}

\begin{thebibliography}{}

\bibitem[\protect\citeauthoryear{Alzantot, Balaji, and
  Srivastava}{2018}]{alzantot2018did}
Alzantot, M.; Balaji, B.; and Srivastava, M.~B.
\newblock 2018.
\newblock Did you hear that? adversarial examples against automatic speech
  recognition.
\newblock {\em CoRR} abs/1801.00554.

\bibitem[\protect\citeauthoryear{Amari and Nagaoka}{2007}]{amari2007methods}
Amari, S.-i., and Nagaoka, H.
\newblock 2007.
\newblock {\em Methods of information geometry}, volume 191.
\newblock American Mathematical Soc.

\bibitem[\protect\citeauthoryear{Amari}{1998}]{amari1998natural}
Amari, S.-I.
\newblock 1998.
\newblock Natural gradient works efficiently in learning.
\newblock {\em Neural Comput.} 10(2):251--276.

\bibitem[\protect\citeauthoryear{Biggio, Fumera, and
  Roli}{2014}]{biggio2014security}
Biggio, B.; Fumera, G.; and Roli, F.
\newblock 2014.
\newblock Security evaluation of pattern classifiers under attack.
\newblock {\em IEEE TKDE} 26(4):984--996.

\bibitem[\protect\citeauthoryear{Brendel, Rauber, and
  Bethge}{2017}]{brendel2017decision}
Brendel, W.; Rauber, J.; and Bethge, M.
\newblock 2017.
\newblock Decision-based adversarial attacks: Reliable attacks against
  black-box machine learning models.
\newblock {\em arXiv preprint arXiv:1712.04248}.

\bibitem[\protect\citeauthoryear{Bul{\`o}, Biggio, and et.
  al.}{2017}]{rota2017randomized}
Bul{\`o}, S.~R.; Biggio, B.; and et. al.
\newblock 2017.
\newblock Randomized prediction games for adversarial machine learning.
\newblock {\em IEEE Transactions on Neural Networks and Learning Systems}
  28(11):2466--2478.

\bibitem[\protect\citeauthoryear{Carlini and Wagner}{2017}]{carlini2017towards}
Carlini, N., and Wagner, D.
\newblock 2017.
\newblock Towards evaluating the robustness of neural networks.
\newblock In {\em Security and Privacy (SP), 2017 IEEE Symposium on},  39--57.
\newblock IEEE.

\bibitem[\protect\citeauthoryear{Chen \bgroup et al\mbox.\egroup
  }{2017a}]{chen2017ead}
Chen, P.-Y.; Sharma, Y.; Zhang, H.; Yi, J.; and Hsieh, C.-J.
\newblock 2017a.
\newblock Ead: elastic-net attacks to deep neural networks via adversarial
  examples.
\newblock {\em arXiv preprint arXiv:1709.04114}.

\bibitem[\protect\citeauthoryear{Chen \bgroup et al\mbox.\egroup
  }{2017b}]{chen2017zoo}
Chen, P.-Y.; Zhang, H.; Sharma, Y.; Yi, J.; and Hsieh, C.-J.
\newblock 2017b.
\newblock Zoo: Zeroth order optimization based black-box attacks to deep neural
  networks without training substitute models.
\newblock In {\em Proceedings of the 10th ACM Workshop on Artificial
  Intelligence and Security},  15--26.
\newblock ACM.

\bibitem[\protect\citeauthoryear{Cox and Hinkley}{1979}]{cox1979theoretical}
Cox, D.~R., and Hinkley, D.~V.
\newblock 1979.
\newblock {\em Theoretical statistics}.
\newblock Chapman and Hall/CRC.

\bibitem[\protect\citeauthoryear{Demontis, Melis, and et.
  al.}{2018}]{demontis2018yes}
Demontis, A.; Melis, M.; and et. al.
\newblock 2018.
\newblock Yes, machine learning can be more secure! a case study on android
  malware detection.
\newblock {\em IEEE TDSC}  1--1.

\bibitem[\protect\citeauthoryear{Deng \bgroup et al\mbox.\egroup
  }{2009}]{deng2009imagenet}
Deng, J.; Dong, W.; Socher, R.; Li, L.-J.; Li, K.; and Fei-Fei, L.
\newblock 2009.
\newblock Imagenet: A large-scale hierarchical image database.
\newblock In {\em Computer Vision and Pattern Recognition, 2009. CVPR 2009.
  IEEE Conference on},  248--255.
\newblock IEEE.

\bibitem[\protect\citeauthoryear{{Duchi} \bgroup et al\mbox.\egroup
  }{2015}]{john2015optimal}
{Duchi}, J.~C.; {Jordan}, M.~I.; {Wainwright}, M.~J.; and {Wibisono}, A.
\newblock 2015.
\newblock Optimal rates for zero-order convex optimization: The power of two
  function evaluations.
\newblock {\em IEEE Transactions on Information Theory} 61(5):2788--2806.

\bibitem[\protect\citeauthoryear{Goodfellow, Shlens, and
  Szegedy}{2015}]{goodfellow2015explaining}
Goodfellow, I.; Shlens, J.; and Szegedy, C.
\newblock 2015.
\newblock Explaining and harnessing adversarial examples.
\newblock {\em 2015 ICLR} arXiv preprint arXiv:1412.6572.

\bibitem[\protect\citeauthoryear{Grosse and
  Martens}{2016}]{grosse2016kronecker}
Grosse, R.~B., and Martens, J.
\newblock 2016.
\newblock A kronecker-factored approximate fisher matrix for convolution
  layers.
\newblock In {\em ICML}, volume~1, ~2.

\bibitem[\protect\citeauthoryear{Grosse and
  Salakhudinov}{2015}]{grosse2015scaling}
Grosse, R., and Salakhudinov, R.
\newblock 2015.
\newblock Scaling up natural gradient by sparsely factorizing the inverse
  fisher matrix.
\newblock In Bach, F., and Blei, D., eds., {\em Proceedings of the 32nd
  International Conference on Machine Learning}, volume~37 of {\em Proceedings
  of Machine Learning Research},  2304--2313.
\newblock Lille, France: PMLR.

\bibitem[\protect\citeauthoryear{Ilyas \bgroup et al\mbox.\egroup
  }{2018}]{ilyas2018blackbox}
Ilyas, A.; Engstrom, L.; Athalye, A.; and Lin, J.
\newblock 2018.
\newblock Black-box adversarial attacks with limited queries and information.
\newblock In {\em Proceedings of the 35th International Conference on Machine
  Learning, {ICML} 2018}.

\bibitem[\protect\citeauthoryear{Ilyas, Engstrom, and
  Madry}{2018}]{ilyas2018prior}
Ilyas, A.; Engstrom, L.; and Madry, A.
\newblock 2018.
\newblock Prior convictions: Black-box adversarial attacks with bandits and
  priors.
\newblock {\em ICLR 2019}.

\bibitem[\protect\citeauthoryear{Karakida, Akaho, and
  Amari}{2018}]{karakida2018universal}
Karakida, R.; Akaho, S.; and Amari, S.-i.
\newblock 2018.
\newblock Universal statistics of fisher information in deep neural networks:
  Mean field approach.
\newblock {\em arXiv preprint arXiv:1806.01316}.

\bibitem[\protect\citeauthoryear{Krizhevsky and
  Hinton}{2009}]{Krizhevsky2009learning}
Krizhevsky, A., and Hinton, G.
\newblock 2009.
\newblock Learning multiple layers of features from tiny images.
\newblock {\em Master's thesis, Department of Computer Science, University of
  Toronto}.

\bibitem[\protect\citeauthoryear{Kullback and
  Leibler}{1951}]{kullback1951information}
Kullback, S., and Leibler, R.~A.
\newblock 1951.
\newblock On information and sufficiency.
\newblock {\em Ann. Math. Statist.} 22(1):79--86.

\bibitem[\protect\citeauthoryear{LeCun, Bengio, and
  Hinton}{2015}]{lecun2015deep}
LeCun, Y.; Bengio, Y.; and Hinton, G.
\newblock 2015.
\newblock Deep learning.
\newblock 521:436--44.

\bibitem[\protect\citeauthoryear{Lecun \bgroup et al\mbox.\egroup
  }{1998}]{Lecun1998gradient}
Lecun, Y.; Bottou, L.; Bengio, Y.; and Haffner, P.
\newblock 1998.
\newblock Gradient-based learning applied to document recognition.
\newblock {\em Proceedings of the IEEE} 86(11):2278--2324.

\bibitem[\protect\citeauthoryear{{Li} \bgroup et al\mbox.\egroup
  }{2019}]{8675229}
{Li}, Z.; {Ding}, C.; {Wang}, S.; {Wen}, W.; {Zhuo}, Y.; {Liu}, C.; {Qiu}, Q.;
  {Xu}, W.; {Lin}, X.; {Qian}, X.; and {Wang}, Y.
\newblock 2019.
\newblock E-rnn: Design optimization for efficient recurrent neural networks in
  fpgas.
\newblock In {\em 2019 IEEE International Symposium on High Performance
  Computer Architecture (HPCA)},  69--80.

\bibitem[\protect\citeauthoryear{Lin, Hong, and et. al.}{2017}]{lin2017tactics}
Lin, Y.-C.; Hong, Z.-W.; and et. al.
\newblock 2017.
\newblock Tactics of adversarial attack on deep reinforcement learning agents.
\newblock In {\em Proceedings of the 26th International Joint Conference on
  Artificial Intelligence},  3756--3762.
\newblock AAAI Press.

\bibitem[\protect\citeauthoryear{Liu \bgroup et al\mbox.\egroup
  }{2019}]{liu2018signsgd}
Liu, S.; Chen, P.-Y.; Chen, X.; and Hong, M.
\newblock 2019.
\newblock {SignSGD} via zeroth-order oracle.
\newblock {\em International Conference on Learning Representations}.

\bibitem[\protect\citeauthoryear{Madry \bgroup et al\mbox.\egroup
  }{2017}]{madry2017towards}
Madry, A.; Makelov, A.; Schmidt, L.; Tsipras, D.; and Vladu, A.
\newblock 2017.
\newblock Towards deep learning models resistant to adversarial attacks.
\newblock {\em arXiv preprint arXiv:1706.06083}.

\bibitem[\protect\citeauthoryear{Martens and
  Grosse}{2015}]{martens2015optimizing}
Martens, J., and Grosse, R.
\newblock 2015.
\newblock Optimizing neural networks with kronecker-factored approximate
  curvature.
\newblock In {\em International conference on machine learning},  2408--2417.

\bibitem[\protect\citeauthoryear{Martens}{2014}]{james2014new}
Martens, J.
\newblock 2014.
\newblock New perspectives on the natural gradient method.
\newblock {\em CoRR} abs/1412.1193.

\bibitem[\protect\citeauthoryear{Martens}{2016}]{martens2016second}
Martens, J.
\newblock 2016.
\newblock {\em Second-order optimization for neural networks}.
\newblock University of Toronto (Canada).

\bibitem[\protect\citeauthoryear{Nesterov and
  Spokoiny}{2017}]{nesterov2017random}
Nesterov, Y., and Spokoiny, V.
\newblock 2017.
\newblock Random gradient-free minimization of convex functions.
\newblock {\em Foundations of Computational Mathematics} 17(2):527--566.

\bibitem[\protect\citeauthoryear{Ollivier}{2015}]{ollivier2015riemannian}
Ollivier, Y.
\newblock 2015.
\newblock Riemannian metrics for neural networks i: feedforward networks.
\newblock {\em Information and Inference: A Journal of the IMA} 4(2):108--153.

\bibitem[\protect\citeauthoryear{Osawa \bgroup et al\mbox.\egroup
  }{2018}]{kazuki2018second}
Osawa, K.; Tsuji, Y.; Ueno, Y.; Naruse, A.; Yokota, R.; and Matsuoka, S.
\newblock 2018.
\newblock Second-order optimization method for large mini-batch: Training
  resnet-50 on imagenet in 35 epochs.
\newblock {\em CVPR 2019}.

\bibitem[\protect\citeauthoryear{Papernot, McDaniel, and
  Goodfellow}{2016}]{nicolas2016transferability}
Papernot, N.; McDaniel, P.~D.; and Goodfellow, I.~J.
\newblock 2016.
\newblock Transferability in machine learning: from phenomena to black-box
  attacks using adversarial samples.
\newblock {\em CoRR} abs/1605.07277.

\bibitem[\protect\citeauthoryear{Pascanu and Bengio}{2013a}]{razvan2013natural}
Pascanu, R., and Bengio, Y.
\newblock 2013a.
\newblock Natural gradient revisited.
\newblock {\em CoRR} abs/1301.3584.

\bibitem[\protect\citeauthoryear{Pascanu and
  Bengio}{2013b}]{pascanu2013revisiting}
Pascanu, R., and Bengio, Y.
\newblock 2013b.
\newblock Revisiting natural gradient for deep networks.
\newblock {\em arXiv preprint arXiv:1301.3584}.

\bibitem[\protect\citeauthoryear{Szegedy \bgroup et al\mbox.\egroup
  }{2016}]{Szegedy2016RethinkingTI}
Szegedy, C.; Vanhoucke, V.; Ioffe, S.; Shlens, J.; and Wojna, Z.
\newblock 2016.
\newblock Rethinking the inception architecture for computer vision.
\newblock {\em CVPR 2016}  2818--2826.

\bibitem[\protect\citeauthoryear{Tu \bgroup et al\mbox.\egroup
  }{2018}]{tu2018autozoom}
Tu, C.-C.; Ting, P.; Chen, P.-Y.; Liu, S.; Zhang, H.; Yi, J.; Hsieh, C.-J.; and
  Cheng, S.-M.
\newblock 2018.
\newblock Autozoom: Autoencoder-based zeroth order optimization method for
  attacking black-box neural networks.
\newblock {\em arXiv preprint arXiv:1805.11770}.

\bibitem[\protect\citeauthoryear{{Wang} \bgroup et al\mbox.\egroup
  }{2018a}]{8646578}
{Wang}, S.; {Wang}, X.; {Ye}, S.; {Zhao}, P.; and {Lin}, X.
\newblock 2018a.
\newblock Defending dnn adversarial attacks with pruning and logits
  augmentation.
\newblock In {\em 2018 IEEE Global Conference on Signal and Information
  Processing (GlobalSIP)},  1144--1148.

\bibitem[\protect\citeauthoryear{Wang \bgroup et al\mbox.\egroup
  }{2018b}]{wang2018defensive}
Wang, S.; Wang, X.; Zhao, P.; Wen, W.; Kaeli, D.; Chin, P.; and Lin, X.
\newblock 2018b.
\newblock Defensive dropout for hardening deep neural networks under
  adversarial attacks.
\newblock In {\em ICCAD '18}.

\bibitem[\protect\citeauthoryear{Wang \bgroup et al\mbox.\egroup
  }{2018c}]{yining2018stochastic}
Wang, Y.; Du, S.; Balakrishnan, S.; and Singh, A.
\newblock 2018c.
\newblock Stochastic zeroth-order optimization in high dimensions.
\newblock In {\em AISTATS 2018}, volume~84 of {\em Proceedings of Machine
  Learning Research}.
\newblock PMLR.

\bibitem[\protect\citeauthoryear{Wang \bgroup et al\mbox.\egroup
  }{2019}]{Wang2019HRS}
Wang, X.; Wang, S.; Chen, P.-Y.; Wang, Y.; Kulis, B.; Lin, X.; and Chin, S.
\newblock 2019.
\newblock Protecting neural networks with hierarchical random switching:
  Towards better robustness-accuracy trade-off for stochastic defenses.
\newblock In {\em IJCAI 2019}.

\bibitem[\protect\citeauthoryear{Xie, Wang, and et.
  al.}{2018}]{xie2017adversarial}
Xie, C.; Wang, J.; and et. al.
\newblock 2018.
\newblock Adversarial examples for semantic segmentation and object detection.
\newblock In {\em ICCV 2017},  1378--1387.

\bibitem[\protect\citeauthoryear{{Xu} \bgroup et al\mbox.\egroup
  }{2018}]{xu2018structured}
{Xu}, K.; {Liu}, S.; {Zhao}, P.; {Chen}, P.-Y.; {Zhang}, H.; {Erdogmus}, D.;
  {Wang}, Y.; and {Lin}, X.
\newblock 2018.
\newblock {Structured Adversarial Attack: Towards General Implementation and
  Better Interpretability}.
\newblock {\em ArXiv e-prints}.

\bibitem[\protect\citeauthoryear{Xu \bgroup et al\mbox.\egroup
  }{2019}]{xu2019topology}
Xu, K.; Chen, H.; Liu, S.; Chen, P.-Y.; Weng, T.-W.; Hong, M.; and Lin, X.
\newblock 2019.
\newblock Topology attack and defense for graph neural networks: An
  optimization perspective.
\newblock {\em arXiv preprint arXiv:1906.04214}.

\bibitem[\protect\citeauthoryear{Ye \bgroup et al\mbox.\egroup
  }{2018}]{ye2018hessianaware}
Ye, H.; Huang, Z.; Fang, C.; Li, C.~J.; and Zhang, T.
\newblock 2018.
\newblock Hessian-aware zeroth-order optimization for black-box adversarial
  attack.
\newblock {\em CoRR} abs/1812.11377.

\bibitem[\protect\citeauthoryear{{Zhang}, {Weng}, and et.
  al.}{2018}]{zhang2018efficient}
{Zhang}, H.; {Weng}, T.-W.; and et. al.
\newblock 2018.
\newblock {Efficient Neural Network Robustness Certification with General
  Activation Functions}.
\newblock In {\em NIPS 2018}.

\bibitem[\protect\citeauthoryear{{Zhao} \bgroup et al\mbox.\egroup
  }{2017}]{7970161}
{Zhao}, A.; {Fu}, K.; {Wang}, S.; {Zuo}, J.; {Zhang}, Y.; {Hu}, Y.; and {Wang},
  H.
\newblock 2017.
\newblock Aircraft recognition based on landmark detection in remote sensing
  images.
\newblock {\em IEEE Geoscience and Remote Sensing Letters} 14(8):1413--1417.

\bibitem[\protect\citeauthoryear{Zhao \bgroup et al\mbox.\egroup
  }{2018}]{zhao2018admm}
Zhao, P.; Liu, S.; Wang, Y.; and Lin, X.
\newblock 2018.
\newblock An admm-based universal framework for adversarial attacks on deep
  neural networks.
\newblock In {\em ACM Multimedia 2018}.

\bibitem[\protect\citeauthoryear{Zhao \bgroup et al\mbox.\egroup
  }{2019a}]{zhao2019design}
Zhao, P.; Liu, S.; Chen, P.-Y.; Hoang, N.; Xu, K.; Kailkhura, B.; and Lin, X.
\newblock 2019a.
\newblock On the design of black-box adversarial examples by leveraging
  gradient-free optimization and operator splitting method.
\newblock In {\em ICCV 2019}.

\bibitem[\protect\citeauthoryear{Zhao \bgroup et al\mbox.\egroup
  }{2019b}]{pu2019fault}
Zhao, P.; Wang, S.; Gongye, C.; Wang, Y.; Fei, Y.; and Lin, X.
\newblock 2019b.
\newblock Fault sneaking attack: A stealthy framework for misleading deep
  neural networks.
\newblock In {\em DAC 2019}.

\bibitem[\protect\citeauthoryear{Zhao \bgroup et al\mbox.\egroup
  }{2019c}]{pu2019admm}
Zhao, P.; Xu, K.; Liu, S.; Wang, Y.; and Lin, X.
\newblock 2019c.
\newblock Admm attack: An enhanced adversarial attack for deep neural networks
  with undetectable distortions.
\newblock In {\em ASPDAC 2019}.

\end{thebibliography}
}

\clearpage
\begin{appendices}
\setcounter{table}{0} 
\renewcommand{\thetable}{A\arabic{table}}
\setcounter{figure}{0} 
\renewcommand\thefigure{A\arabic{figure}} 

\section{Appendix}

\section{Geometric Interpretation} \label{Ageometric_interpretation}
We provide a geometric interpretation for the natural gradient in this section.
The negative gradient $- \nabla f (\bm \delta)$ can be interpreted as the steepest descent direction  in the sense that it yields the most reduction in  $f$ per unit of change of $\bm \delta$, where the change is measured by the standard Euclidean norm $\left\|  \cdot  \right\|$ \cite{james2014new}, as shown below,
{\small\begin{equation}
\frac{{ - \nabla f(\bm \delta)}}{{\left\| {\nabla f(\bm \delta)} \right\|}} = \mathop {\lim }\limits_{\epsilon  \to 0} \frac{1}{\epsilon }\arg \mathop {\min }\limits_{\left\| \alpha  \right\| \le \epsilon } f(\bm \delta  + \bm \alpha ).
\end{equation}}%
By  following the $- \nabla f(\bm \delta)$ direction, we can obtain the change of $\bm \delta$ within a certain $\epsilon$-neighbourhood to minimize the loss function.

As the loss function is related to the likelihood, we can explore the steepest direction to minimize the loss in the space of all possible likelihoods (i.e. distribution space). 
KL divergence \cite{kullback1951information} is a popular measure of the distance between two distributions. %although it is only approximately symmetric within a local neighbourhood. 
For two distributions $p\left( t|\bm{\delta } \right)$ and $p\left( t|\bm{\delta }' \right)$, their KL divergence is defined as 
{\small\begin{equation}
\text{KL}\left[ p\left( t|\bm{\delta } \right) ||p\left( t|\bm{\delta }' \right) \right] =\int{p\left( t|\bm{\delta } \right) \log \frac{p\left( t|\bm{\delta } \right)}{p\left( t|\bm{\delta }' \right)}}dt.
\end{equation}}%

\begin{lemma} \label{KL-Hessian}
The FIM $\bm F $ is the Hessian of KL divergence between two distributions $p(t|\bm \delta)$ and $p(t|\bm \delta + \bm \alpha)$, with respect to $\bm \alpha$, evaluated at $\bm \alpha =\bm 0$.
\end{lemma}
The proof is shown in the appendix. By Lemma \ref{KL-Hessian}, the FIM can be regarded as the curvature in the distribution space.

\begin{lemma} \label{KL-second-order-expansion}
The second order Taylor expansion of the KL divergence can be expressed as 
{ \small \begin{equation} 
 {\rm{KL}}\left[ {p(t|{\bm{\delta }})||p(t|{\bm{\delta }} + {\bm{\alpha }})} \right] \approx \frac{1}{2}{{\bm{\alpha }}^{\rm{T}}}{\bf{F \bm \alpha }}.
 \end{equation}}%
\end{lemma}
We provide a proof in the appendix. 

Next we explore the direction to minimize the loss function in the distribution space where the distance is measured by the KL divergence. Although in general, the KL divergence is not symmetric, it is (approximately) symmetric in a local neighborhood. The problem can be formulated as
{\small\begin{equation}
{{\bm{\alpha }}^*} = \mathop {\arg \min }\limits_{{\rm{KL}}\left[ {p(t|{\bm{\delta }})||p(t|{\bm{\delta }} + {\bm{\alpha }})} \right] = m} f({\bm{\delta }} + {\bm{\alpha }}).
\end{equation}}%
where $m$ is a certain constant. The purpose of fixing the KL divergence to a constant is to  move along the distribution space with a constant speed, regardless of the curvature.

%\begin{lemma}  \label{NG-distribution2}
%The negative natural gradient is the steepest descent direction in the distribution space.
%\end{lemma}

Thus, we can obtain Lemma  \ref{NG-distribution} and show  its proof in the appendix. In the parameter space, the negative gradient is the steepest descent direction to minimize the loss function. By contrast, in the distribution space, the steepest descent direction  is  the negative  natural gradient. Thus, the direction in distribution space defined  by the natural gradient will be invariant to the choice of parameterization \cite{pascanu2013revisiting}, i.e., it will not be affected by how the model is parametrized, but only depends on the distribution induced by the parameters.

\section{Proof of Lemmas}

\subsection{Proof of Lemma \ref{score-expectation}}

\textit{Proof of Lemma \ref{score-expectation}.}
\begin{align}
\mathop {\mathop{\mathbb E}\nolimits} \limits_{p(t, \bm x| \bm \delta)} \left[ {s({\bm{\delta }})} \right] & =  \mathop {\mathop{\mathbb E}\nolimits} \limits_{p(t, \bm x| \bm \delta)} \left[ {\nabla \log p(t, \bm x| \bm \delta)} \right] \nonumber \\
& =  \int {\nabla \log p(t, \bm x| \bm \delta)} p(t, \bm x| \bm \delta) dt \ d{\bm{x }} \nonumber \\
& = \int {\frac{{\nabla p(t, \bm x| \bm \delta)}}{{p(t, \bm x| \bm \delta)}}} p(t, \bm x| \bm \delta)dt \ d{\bm{x }} \nonumber \\
& = \nabla \int p(t, \bm x| \bm \delta) dt \ d{\bm{x }} \nonumber \\
& = 0 \nonumber
\end{align}

\subsection{Proof of Lemma \ref{KL-Hessian}}

\textit{Proof of Lemma \ref{KL-Hessian}.}
The gradients of the KL divergence can be expressed as
\begin{align}
{\nabla _{\bf{\alpha }}}{\rm{KL}}\left[ {p(t|{\bf{\delta }})||p(t|{\bf{\delta }} + {\bf{\alpha }})} \right] &= {\nabla _{\bf{\alpha }}}\int {p(t|{\bf{\delta }})\log \frac{{p(t|{\bf{\delta }})}}{{p(t|{\bf{\delta }} + {\bf{\alpha }})}}} dt \nonumber \\
 &= {\nabla _{\bf{\alpha }}}\int {p(t|{\bf{\delta }})\log p(t|{\bf{\delta }})} dt  \nonumber \\
 & \ \ \ \  - {\nabla _{\bf{\alpha }}}\int {p(t|{\bf{\delta }})\log p(t|{\bf{\delta }} + {\bf{\alpha }})} dt\nonumber  \\
& =  - {\nabla _{\bf{\alpha }}}\int {p(t|{\bf{\delta }})\log p(t|{\bf{\delta }} + {\bf{\alpha }})} dt \nonumber  \\
& =  - \int {p(t|{\bf{\delta }}) {\nabla _{\bf{\alpha }}}\log p(t|{\bf{\delta }} + {\bf{\alpha }})} dt
\end{align}

\begin{equation}
\nabla _{\bf{\alpha }}^2{\rm{KL}}\left[ {p(t|{\bf{\delta }})||p(t|{\bf{\delta }} + {\bf{\alpha }})} \right] =  - \int {p(t|{\bf{\delta }})\nabla _{\bf{\alpha }}^2\log p(t|{\bf{\delta }} + {\bf{\alpha }})} dt
\end{equation}
The Hessian of the KL divergence is defined by
\begin{align}
{{\bf{H}}_{{\rm{KL}}\left[ {p(t|{\bf{\delta }})||p(t|{\bf{\delta }} + {\bf{\alpha }})} \right]}} &=  - \int {p(t|{\bf{\delta }})\nabla _{\bf{\alpha }}^2\log p(t|{\bf{\delta }} + {\bf{\alpha }}){|_{{\bf{\alpha }} = {\bf{0}}}}} dt  \nonumber  \\
 &=  - \int {p(t|{\bf{\delta }}){{\bf{H}}_{\log p(t|{\bf{\delta }})}}} dt  \nonumber  \\
& =  - \mathop E\limits_{p(t|{\bf{\delta }})} \left[ {{{\bf{H}}_{\log p(t|{\bf{\delta }})}}} \right] \nonumber  \\
 &= {\bf{F}}
\end{align}

\subsection{Proof of Lemma \ref{KL-second-order-expansion}}
\textit{Proof of Lemma \ref{KL-second-order-expansion}.}
\begin{align}
{\rm{KL}}\left[ {p(t|{\bf{\delta }})||p(t|{\bf{\delta }} 
 + {\bf{\alpha }})} \right]
 & \approx {\rm{KL}}\left[ {p(t|{\bf{\delta }})||p(t|{\bf{\delta }})} \right] \nonumber \\
 &  \ \ \ \ + {\left( {\nabla _{\bf{\alpha }}^{}{\rm{KL}}\left[ {p(t|{\bf{\delta }})||p(t|{\bf{\delta }}   
 + {\bf{\alpha }})} \right]{|_{{\bf{\alpha }} 
= {\bf{0}}}}} \right)^{\rm{T}}}{\bf{\alpha }} \nonumber \\
 & \ \ \ \ + \frac{1}{2}{{\bf{\alpha }}^{\rm{T}}}{\bf{F\alpha }} \nonumber \\
& = {\rm{KL}}\left[ {p(t|{\bf{\delta }})||p(t|{\bf{\delta }})} \right] \nonumber \\
& \ \ \ \ - \mathop E\limits_{p(t|{\bf{\delta }})} {\left[ {\nabla _{\bf{\delta }}^{}\log p(t|{\bf{\delta }})} \right]^{\rm{T}}}{\bf{\alpha }}  \nonumber \\
& \ \ \ \ + \frac{1}{2}{{\bf{\alpha }}^{\rm{T}}}{\bf{F\alpha }}
\end{align}
Notice that the first term is zero as they are the same distributions. The second term is zero due to Lemma \ref{score-expectation}.

\subsection{Proof of Lemma \ref{NG-distribution}}
\textit{Proof of Lemma \ref{NG-distribution}.}
The Lagrangian function of the minimization can be formulated as
\begin{align}
L({\bf{\alpha }}) & = f({\bf{\delta }} + {\bf{\alpha }}) + \phi \left( {{\rm{KL}}\left[ {p(t|{\bf{\delta }})||p(t|{\bf{\delta }} + {\bf{\alpha }})} \right] - c} \right) \nonumber \\
 & \approx f({\bf{\delta }}) + \nabla f{({\bf{\delta }})^{\rm{T}}}{\bf{\alpha }} + \frac{1}{2}\phi {{\bf{\alpha }}^{\rm{T}}}{\bf{F\alpha }} - \phi c
\end{align}
To solve this minimization, we set its derivative to zero:
\begin{equation}
\frac{\partial }{{\partial {\bf{\alpha }}}}\left( {f({\bf{\delta }}) + {\nabla _{\bf{\delta }}}f{{({\bf{\delta }})}^{\rm{T}}}{\bf{\alpha }} + \frac{1}{2}\varphi {{\bf{\alpha }}^{\rm{T}}}{\bf{F\alpha }} - \varphi c} \right) = 0
\end{equation}
\begin{equation}
{\nabla _{\bf{\delta }}}f({\bf{\delta }}) + \varphi {\bf{F\alpha }} = 0
\end{equation}
\begin{equation}
{\bf{\alpha }} =  - \frac{1}{\varphi }{{\bf{F}}^{ - 1}}{\nabla _{\bf{\delta }}}f({\bf{\delta }})
\end{equation}
We can see that the negative natural gradient defines the steepest direction in the distribution space.

\section{Adversarial Examples}

\begin{figure}[h]    
 \centering
\begin{tabular}{p{0.1in}p{0.4in}p{0.4in}p{0.4in}p{0.4in}p{0.4in}}
& \parbox{0.5in}{\centering \footnotesize 3} &  
\parbox{0.5in}{\centering \footnotesize 6}
&  
\parbox{0.5in}{\centering \footnotesize 8  }
&\parbox{0.5in}{\centering \footnotesize 7  }
&\parbox{0.5in}{\centering \footnotesize 5  }
\\
 \rotatebox{90}{\parbox{0.5in}{\centering \footnotesize MNIST }}
 &   %\hspace*{-0.1in} 
\includegraphics[width=0.5in]{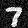}& %\hspace*{-0.1in}  
\includegraphics[width=0.5in]{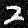}& %\hspace*{-0.1in}
\includegraphics[width=0.5in]{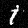}&
\includegraphics[width=0.5in]{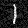}&
\includegraphics[width=0.5in]{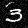}
\\ 
& \parbox{0.5in}{\centering \footnotesize deer} &  
\parbox{0.5in}{\centering \footnotesize ship}
&  
\parbox{0.5in}{\centering \footnotesize truck  }
&  
\parbox{0.5in}{\centering \footnotesize frog  }
&  
\parbox{0.5in}{\centering \footnotesize cat  }
\\
%\rotatebox{90}{ \footnotesize \ \ \ \ Adversarial}
 \rotatebox{90}{\parbox{0.5in}{\centering \footnotesize CIFAR }}
&   %\hspace*{-0.1in}
\includegraphics[width=0.5in]{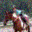} &  %\hspace*{-0.1in}
\includegraphics[width=0.5in]{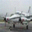} &  %\hspace*{-0.1in}
\includegraphics[width=0.5in]{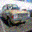} &
\includegraphics[width=0.5in]{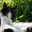} &
\includegraphics[width=0.5in]{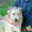} 
\end{tabular}
\caption{\footnotesize{ The legitimate images and their adversarial examples generated by ZO-NGD on MNIST and CIFAR-10. The misclassified class are shown on top of each image.
}} \label{fig: examples_MNIST_CIFAR}
\end{figure}

\section{Ablation study }

\subsubsection{Various Models}
To check the performance of the proposed method on various model architecture, we performed  experiments on three datasets  using two new models (VGG16 \& ResNet) and summarized the results  in Table\,\ref{tab: models}. The proposed ZO-NGD is more query-efficient than NES-PGD. 

\begin{table}[htb]
\centering
\caption{ Attack success rate (ASR) and query \# for two new models.}
\scalebox{1.1}[1.1]{
\begin{threeparttable}
\label{tab: models}
{\small
\begin{tabular}{c|c|c|c|c}
\toprule%
 & \multicolumn{2}{c|}{ MNIST (VGG)} & \multicolumn{2}{c}{ CIFAR10 (VGG)}  \\
\hline
& ASR &   Query  & ASR & Query      \\
\hline
NES-PGD & 99.2\% & 1082 &  98.3\% &  381   \\
ZO-NGD  &  99.5\% & 548  &  98.2\% & 152  \\
\midrule
\midrule
   & \multicolumn{2}{c|}{ ImageNet (VGG) }  & \multicolumn{2}{c}{ ImageNet (ResNet) } \\
\hline
& ASR &   Query  & ASR & Query       \\
\hline
NES-PGD  & 96.8\% & 1136 & 97.2\%  & 1281 \\
ZO-NGD    & 96.5\% & 594 & 98.1\%  & 624 \\
\bottomrule
\end{tabular}}
%\begin{tablenotes}
%\end{tablenotes}
\end{threeparttable}}
\end{table}

%\vspace{-0.3cm}
\iffalse
\begin{table}[htb]
\centering
\caption{ Attack success rate (ASR) and query \# for two new models.}
\scalebox{0.65}[0.65]{
\begin{threeparttable}
\label{tab: models}
{\small
\begin{tabular}{c|c|c|c|c|c|c|c|c}
\hline%
 & \multicolumn{2}{c|}{ MNIST (VGG)} & \multicolumn{2}{c|}{ CIFAR10 (VGG)}  & \multicolumn{2}{c|}{ ImageNet (VGG) }  & \multicolumn{2}{c}{ ImageNet (ResNet) } \\
\hline
& ASR &   Query  & ASR & Query    & ASR &  Query  & ASR &  Query   \\
\hline
NES-PGD & 99.2\% & 1082 &  98.3\% &  381  & 96.8\% & 1136 & 97.2\%  & 1281 \\
ZO-NGD  &  99.5\% & 548  &  98.2\% & 152  & 96.5\% & 594 & 98.1\%  & 624 \\
\hline
\end{tabular}}
%\begin{tablenotes}
%\end{tablenotes}
\end{threeparttable}}
\end{table}
\fi

\subsubsection{Parameter Analysis}
The sensitivity analysis on the parameters $\mu$ and $\epsilon$ are demonstrated in Table\,\ref{tab: parameters}. As observed form Table \ref{tab: parameters}, the ZO-NGD performance is robust to different $\mu$ values (fixing $\epsilon=0.2$), and larger $\epsilon$ leads to fewer queries. 
%\vspace{-0.7cm}
\begin{table}[htb]
\centering
\caption{ Attack success rate (ASR) and query \# for various hyper-parameters on ImageNet (Inception).}
\scalebox{1.1}[1.1]{
\begin{threeparttable}
\label{tab: parameters}
{\small
\begin{tabular}{c|c|c|c|c|c}
\toprule%
  \multicolumn{6}{c}{ { Value of $\mu$, while fixing $\epsilon=0.2$ }}   \\
\hline
  \multicolumn{2}{c|}{ 1} & \multicolumn{2}{c|}{ 0.1}  & \multicolumn{2}{c}{ 0.01 } \\
\hline
 ASR &   Query  & ASR & Query   & ASR &  Query   \\
\hline
97.3\% &  626 & 97\% & 582 & 96.6\% & 596  \\
\midrule
\midrule
 \multicolumn{6}{c}{  { Value of  $\epsilon$, while fixing $\mu=0.1$ } } \\
 \hline
  \multicolumn{2}{c|}{ 0.15 }  & \multicolumn{2}{c|}{ 0.2 }  & \multicolumn{2}{c}{ 0.25 } \\
  \hline
   ASR &  Query  & ASR &  Query  & ASR &  Query \\  \hline
  96.1\%  & 619 & 97\% & 583  & 98.2\% & 559 \\
  \bottomrule
\end{tabular}}
%\begin{tablenotes}
%\end{tablenotes}
\end{threeparttable}}
\end{table}

\iffalse
\begin{table}[htb]
\centering
\scalebox{0.57}[0.57]{
\begin{threeparttable}
\caption{ Attack success rate (ASR) and query \# for various hyper-parameters on ImageNet (Inception).}
\label{tab: parameters}
{\small
\begin{tabular}{c|c|c|c|c|c|c|c|c|c|c|c}
\hline%
  \multicolumn{6}{c|}{ { Value of $\mu$, while fixing $\epsilon=0.2$ }}  & \multicolumn{6}{c}{  { Value of  $\epsilon$, while fixing $\mu=0.1$ } } \\
\hline
  \multicolumn{2}{c|}{ 1} & \multicolumn{2}{c|}{ 0.1}  & \multicolumn{2}{c|}{ 0.01 }  & \multicolumn{2}{c|}{ 0.15 }  & \multicolumn{2}{c|}{ 0.2 }  & \multicolumn{2}{c}{ 0.25 } \\
\hline
 ASR &   Query  & ASR & Query   & ASR &  Query  & ASR &  Query  & ASR &  Query  & ASR &  Query   \\
\hline
97.3\% &  626 & 97\% & 582 & 96.6\% & 596  & 96.1\%  & 619 & 97\% & 583  & 98.2\% & 559 \\
\hline
\end{tabular}}
%\begin{tablenotes}
%\end{tablenotes}
\end{threeparttable}}
\end{table}
\fi

In second-order optimization, damping is a common technique to compensate for errors in the quadratic approximation. The parameter $\gamma$ plays a key role in damping. To show the influence of $\gamma$,  we demonstrate the loss after 2 ADMM iterations for a wide range  of  $\gamma$ values  with  the  same  initialization  in  Figure \ref{fig: pro_distribution}(a).   We  observe  that  0.01  or  0.001  is  an  appropriate choice for  $\gamma$ to achieve higher query efficiency.

%\vspace{-4mm}

%\vspace{-4mm}

\subsubsection{Drift of Outer Product}
\begin{figure}[ht]    
 \centering
\begin{tabular}{p{1.6in}p{1.6in}}
\includegraphics[width=1.46in]{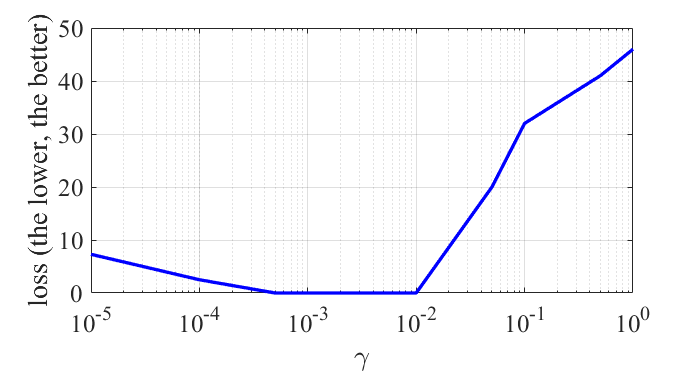} &
\includegraphics[width=1.46in]{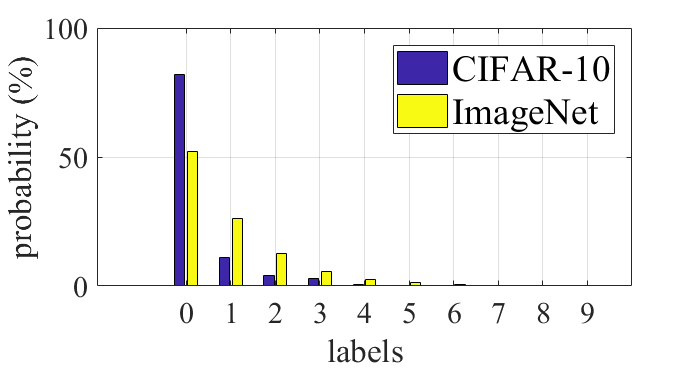} \\
%\vspace{-3mm}
\parbox{1.5in}{\centering \footnotesize (a) Influence of $\gamma$ on MNIST} & 
%\vspace{-3mm}  
\parbox{1.5in}{\centering \footnotesize (b) probability distribution}
\end{tabular}
%\vspace{-2mm} 
\caption{\footnotesize{ Influence of $\gamma$ and prediction probability distribution.
Prediction probability distribution on CIFAR-10/ImageNet.
}} \label{fig: pro_distribution}
\end{figure}

Although we adopt outer product (Equ. \eqref{equ: out_product}) to approximate Equ. \eqref{equ: precise_Fish}, we use the empirical evidence below to motivate why Equ. \eqref{equ: out_product} dominates in Equ.  \eqref{equ: precise_Fish} and the approximation is reasonable. For a well-trained model, the prediction probability of a correctly classified image usually dominates the probability distribution, that is, $p(t|\bm{x},\bm{\delta})$ is usually much larger than other probabilities if $t$ is  correct and $\bm{\delta}$ is small. We plot the average prediction probability distribution of 1000 correctly classified images on CIFAR-10 and ImageNet for their top-10 labels in Figure \ref{fig: pro_distribution}(b). As  observed from Figure \ref{fig: pro_distribution}(b), the correct label usually dominates in the probability distribution, leading to reasonable approximation loss from Equ.  \eqref{equ: precise_Fish} to Equ. \eqref{equ: out_product}.

%\subsubsection{Query Number Distribution} 
%Figure \ref{fig: query_distribution} shows the cumulative distribution of the query number for 1000 images on three datasets.  (XXX figure will be fixed to remove ASR)
%\begin{figure}[h]
%\vspace{-0.4cm}
%\centering
%  \includegraphics[width=0.63\linewidth]{figs/query_distribution_2.png}
  %\vspace{-1mm}
%  \caption{\footnotesize{CDF of query number on three datasets.}} 
%  \label{fig: query_distribution}
  %\vspace{-0.4cm}
%\end{figure}

 \end{appendices}

\end{document}